\documentclass[10pt,twocolumn,letterpaper]{article}

\usepackage{cvpr}
\usepackage{times}
\usepackage{epsfig}
\usepackage{graphicx}
\usepackage{amsmath}
\usepackage{amssymb}

\usepackage[pagebackref=true,breaklinks=true,letterpaper=true,colorlinks,bookmarks=false]{hyperref}

\usepackage{cite}
\usepackage{url}
\usepackage{booktabs}
\usepackage{tabularx}
\usepackage{multirow}
\usepackage{subfigure}
\usepackage{amsmath,dsfont}
\usepackage{array}

\usepackage{algorithm} 
\usepackage{algorithmic} 

\usepackage{pifont}

\usepackage{verbatim}

\usepackage{colortbl}
\definecolor{mygray}{gray}{.75}

\usepackage{booktabs}

\usepackage{xcolor}

\newcommand{\cs}[1]{\textcolor{red}{[\textbf{CS}: #1]}}

\usepackage[normalem]{ulem}
\newcommand{\ourmodel}{Word2VisualVec}

\newcommand{\wordvec}{word2vec}
\newcommand{\videovec}{Word2VideoVec}

\newcommand{\ourmodelbow}{\textit{Word2VisualVec}(\textit{bow})}

\newcommand{\ourmodelwv}{\textit{Word2VisualVec}(\textit{w2v})}

\newcommand{\videovecbow}{\textit{Word2VideoVec}(\textit{bow})}

\newcommand{\videovecwv}{\textit{Word2VideoVec}(\textit{w2v})}

\newcommand{\argmin}{\operatornamewithlimits{argmin}}

\newcommand{\PreserveBackslash}[1]{\let\temp=\\#1\let\\=\temp}
\newcolumntype{C}[1]{>{\PreserveBackslash\centering}p{#1}}
\newcolumntype{R}[1]{>{\PreserveBackslash\raggedleft}p{#1}}
\newcolumntype{L}[1]{>{\PreserveBackslash\raggedright}p{#1}}

\newfont{\mycrnotice}{ptmr8t at 7pt}
\newfont{\myconfname}{ptmri8t at 7pt}
 \cvprfinalcopy 

\def\cvprPaperID{XXXX} 

\ifcvprfinal\fi
\begin{document}

\title{Word2VisualVec: Image and Video to Sentence Matching \\by Visual Feature Prediction}

\author{Jianfeng Dong\\
Zhejiang University\\
{\tt\small danieljf24@zju.edu.cn}
\and
Xirong Li*\\
Renmin University of China\\
{\tt\small xirong@ruc.edu.cn}
\and
Cees G. M. Snoek\\
University of Amsterdam\\
{\tt\small cgmsnoek@uva.nl}
}

\maketitle

\begin{abstract}
This paper strives to find the sentence best describing the content of an image or video. Different from existing works, which rely on a joint subspace for image / video to sentence matching, we propose to do so in a visual space only. We contribute \emph{Word2VisualVec}, a deep neural network architecture that learns to predict a deep visual encoding of textual input based on sentence vectorization and a multi-layer perceptron. We thoroughly analyze its architectural design, by varying the sentence vectorization strategy, network depth and the deep feature to predict for image to sentence matching. We also generalize Word2VisualVec for matching a video to a sentence, by extending the predictive abilities to 3-D ConvNet features as well as a visual-audio representation. Experiments on four challenging image and video benchmarks detail Word2VisualVec's properties, capabilities for image and video to sentence matching, and on all datasets its state-of-the-art results. 
\end{abstract}
%



\section{Introduction} \label{sec:intro}

Given an image or a video, this paper attacks the problem of finding the sentence best describing its content. Since vision and language are two distinct modalities, one has to represent both in a common space wherein the relevance between the two modalities can be computed~\cite{flickr8k,aaai2015-xu-video}. Different from existing approaches for image / video to sentence matching, which rely on a latent subspace \cite{eccv2014-sentence-embedding,cvpr2015-neuraltalk,nips13devise,iccv15-huawei,pan2016jointly}, we propose to match an image or a video to its most likely sentence in a visual space exclusively.

From the vision side we are inspired by the astonishing success of deep convolutional neural networks in image classification \cite{nips2012-hinton,Jia2014Caffe,simonyan2014very,googlenet,cvpr2016-resnet}. These neural networks learn a textual class prediction for an image by successive layers of convolutions, non-linearities, pooling, and full connections, with the aid of big amounts of labeled images, \eg, ImageNet \cite{ILSVRCarxiv14}. Apart from classification, the features derived from the layers of these networks are superior representations for various computer vision challenges  \cite{cvpr2014-cnnbaseline,overfeat}.
We also rely on a layered neural network architecture, but rather than predicting a class label for an image, we strive to predict a deep visual feature from a natural language description for the purpose of matching an image, or a video, to its most likely sentence.

\begin{figure}[t]
\centering
\includegraphics[width=0.99\linewidth]{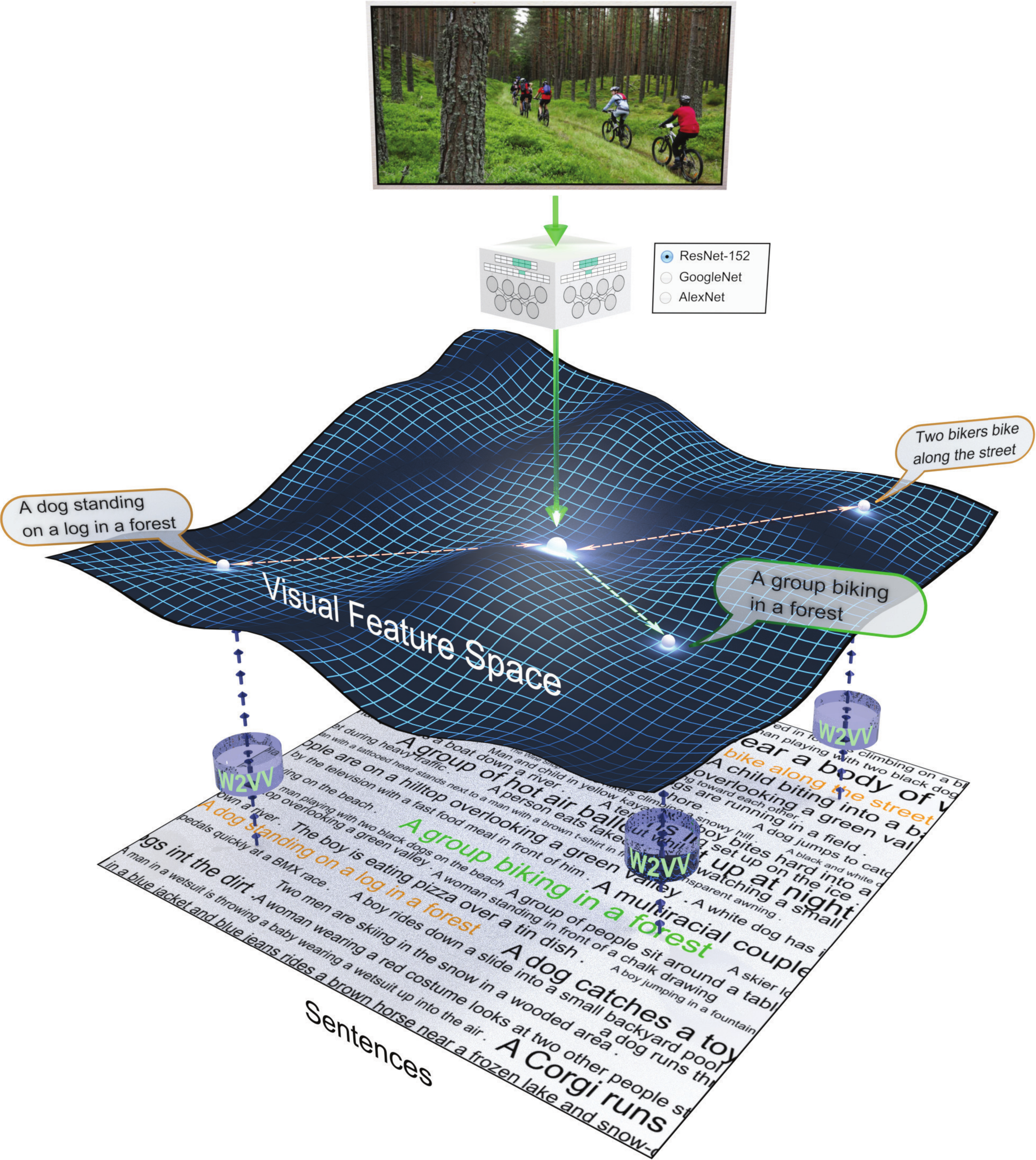}
\caption{\textbf{We propose \emph{Word2VisualVec}}, a deep neural network architecture that strives to find the best sentence describing the content of an image or video. Different from existing works that rely on a latent subspace, we propose to perform the image / video to sentence matching in a visual space only.}
\label{fig:what}
\end{figure}

From the language side we are inspired by the encouraging progress in text vectorization by neural language modeling \cite{word2vec,vinyals2015show,huang2013learning}.
In particular, \wordvec~\cite{word2vec} pre-trained on large-scale text corpora provides distributed word embeddings, an important prerequisite for vectorizing sentences towards a representation shared with image \cite{nips13devise,cvpr2015-klein-fv} or video  \cite{aaai2015-xu-video,JainICCV15}.
Similar to Frome \etal \cite{nips13devise}, we build our model on top of \wordvec~to allow for the handling of a large vocabulary. What is different is that we continue to transform the text embedding from word2vec into a higher-dimensional visual feature space via a multi-layer perceptron. Hence, we predict visual features from text. We call our approach \emph{Word2VisualVec}.

We make three contributions in this paper. First, to the best of our knowledge we are the first to match images to sentences in the visual space only. Second, we propose Word2VisualVec, a deep neural network architecture that learns to predict a deep visual representation of textual sentence input based on sentence vectorization and a multi-layer perceptron. We consider prediction of several recent visual features~\cite{Jia2014Caffe,googlenet,cvpr2016-resnet} based on text, but the approach is general and can, in principle, predict any deep visual feature it is trained on. Third, we generalize Word2VisualVec to match a video to a sentence, by predicting 3-D convolutional neural network features~\cite{iccv2015-c3d} as well as a visual-audio representation including Mel Frequency Cepstral Coefficients~\cite{opensmile}. Experiments on Flickr8k~\cite{flickr8k}, Flickr30k~\cite{flickr30k}, the Microsoft Video Description dataset~\cite{chen2011collecting} and the very recent NIST TrecVid Video-to-Text challenge~\cite{AwadTRECVID16} detail Word2VisualVec's properties, potential for text retrieval, abilities for image and video to sentence matching, and its state-of-the-art results. 
Before detailing our approach, we first highlight in more detail related work.

\section{Related Work} \label{sec:relwork}

\subsection{Sentence vectorization} 
For matching an image or a video with variably-sized sentences, transforming a specific sentence into a fixed-length vector is a prerequisite. The classical bag-of-words representation is adopted in Ba \etal \cite{lei2015predicting}, which is then projected into a 50-dimensional subspace by a multi-layer perceptron (MLP). Fang \etal \cite{fang2015captions} employ a word hashing technique \cite{huang2013learning} to vectorize a sentence before feeding it into a deep multi-modal similarity model. Each word is first decomposed into a list of letter-trigrams, \eg, dog as \{\#do, dog, og\#\}. Consequently, a sentence is represented by a letter-trigram counting vector. Since the number of unique letter-trigrams is less than the number of English words, it has a better scalability than bag of words. Given a sentence vector, at least one-layer embedding is required to place the vector into the shared space, let it be varied subspaces previously exploited or the visual space advocated by this work.

Even with the word hashing tactic, training an embedding matrix for a large vocabulary would require a considerable amount of image-sentence pairs, let alone learning deep embedding models. To conquer the challenge, a distributional text embedding provided by \wordvec~\cite{word2vec} is gaining increased attention. The word embedding matrix used in \cite{nips13devise,cvpr2015-neuraltalk,iclr15-ma-mrnn,aaai2015-xu-video,eccv2016ws-otani} is instantiated by a \wordvec~model pre-trained on large-scale text corpora. In Frome \etal \cite{nips13devise}, for instance, the input text is vectorized by mean pooling the \wordvec~vectors of its words. 
%
%
To better capture visual relationships between words, social tags of Flickr images are used as training data as an alternative to text corpora in \cite{li2015zero}, while Kottur \etal \cite{kottur2015visual} propose an MLP to learn visually grounded \wordvec~from abstract scenes and associated descriptions. As these visual variants of \wordvec~are also to embed words into a hidden space, they are not directly applicable for image or video to sentence matching.

Our proposed \ourmodel~is agnostic to the underlying sentence vectorization strategy and can flexibly embrace bag-of-words, hashing, or word2vec as its input layer.

\subsection{Matching images and videos to sentences}
Prior to deep visual features, works often resort to relatively complicated models to learn a shared representation, in order to compensate the deficiency of traditional low-level visual features. 
Hodosh \etal \cite{flickr8k} leverage Kernel CCA, finding a joint embedding by maximizing the correlation between the projected image and text kernel matrices.
With deep visual features, we observe an increased use of relatively light embeddings on the image side.
Using the fc6 layer of a pre-trained AlexNet \cite{nips2012-hinton} as the image feature, Gong \etal show that linear CCA compares favorably to its kernel counterpart \cite{eccv2014-sentence-embedding}. 
Recent models, \eg, \cite{cvpr2015-neuraltalk,vinyals2015show,iclr15-ma-mrnn},
utilize affine transformation to reduce the image feature to a much shorter $h$-dimensional vector,
with the transformation optimized in an end-to-end fashion within a deep language modeling framework. 
As for the language model, Long Short Term Memory (LSTM) is considered by Vinyals \etal  \cite{vinyals2015show}.
Karpathy and Fei-Fei \cite{cvpr2015-neuraltalk} develop a bi-directional Recurrent Neural Network (RNN) to transform words in a sentence from their original embeddings also to a $h$-dimensional vector.
Mao \etal \cite{iclr15-ma-mrnn} add one more transformation layer before feeding the word vector into RNN.
Although in principle one might consider directly feeding the visual feature into an LSTM or RNN module,
this would let trainable parameters increase in a quadratic order. 
%

\begin{figure*}[t]
\centering
\includegraphics[width=0.97\linewidth]{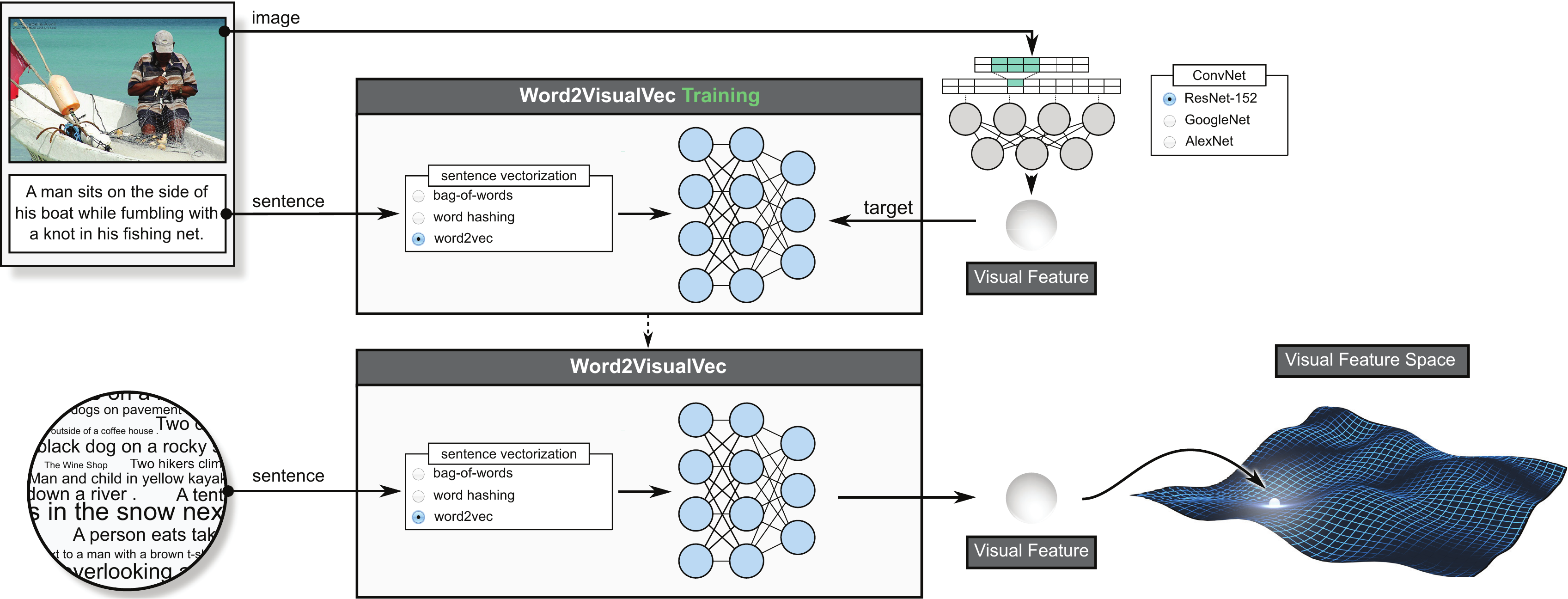}
\caption{\textbf{Word2VisualVec network architecture}. The model first vectorizes an input sentence into a fixed-length vector by relying on bag-of-words, hashing, or word2vec. The vector then goes through a multi-layer perceptron to produce the visual feature vector of choice, for example a GoogleNet or a ResNet. The network parameters are learned from many image-sentence pairs in an end-to-end fashion, with the goal of reconstructing from the input sentence the visual feature vector of the image it is describing. }
\label{fig:how}
\end{figure*}

Similar to the image domain, the state-of-the-art video-to-sentence models are also operated in a shared subspace \cite{deep-video-survey}.
Xu \etal \cite{aaai2015-xu-video} vectorize each subjective-verb-object (svo) triplet extracted from a given sentence by a pre-trained \wordvec, 
and subsequently aggregate the svo vectors into a sentence-level vector by a recursive neural network. 
A joint embedding model is then used to project both the sentence vector and the video vector, obtained by mean pooling over frame-level features, into a latent subspace.
Otani \etal \cite{eccv2016ws-otani} improve upon \cite{aaai2015-xu-video} by exploiting web image search results of an input sentence, which are deemed helpful for word disambiguation, \eg, telling if the word ``keyboard'' refers to a musical instrument or an input device for computers.
Venugopalan \etal  utilize a stacked LSTM to associate a sequence of video frames to a sequence of words \cite{venugopalan2015sequence}.
Yao \etal improve video embedding by relying on a soft attention mechanism to assign larger pooling weights to more important frames \cite{yao2015describing}.
The LSTM-E model by Pan \etal \cite{pan2016jointly} jointly minimizes the existing loss in an LSTM framework and another loss that reflects the distance between the video and sentence embedding vectors in the shared subspace.

The success of deep ConvNet features lets us hypothesize that the deep visual feature space is already good for image / video to sentence comparison and there is no need for further re-projection. Therefore, different from all the above works, we do not impose any projection on the visual feature. Instead, we project the vectorized sentence into the visual space and perform the matching there.



%
%


\section{Word2VisualVec} \label{sec:approach}

Our goal is to learn a visual representation from a natural language description. By doing so, the relevance between a given image $x$ and a specific sentence $q$ can be directly computed in a visual feature space. More formally, let $\phi(x) \in \mathbb{R}^d$ be a $d$-dimensional visual feature vector. A pretrained ConvNet, apart from its original mission of visual class recognition, has now been recognized as an effective visual feature extractor \cite{cvpr2014-cnnbaseline}. We follow this good practice, instantiating $\phi(x)$ with a ConvNet feature vector. We aim for a sentence representation $r(q)\in \mathbb{R}^d$ such that the similarity can be expressed in terms of $\phi(x)$ and $r(q)$, say, in the form of an inner product.
\ourmodel~is designed to produce $r(q)$, as visualized in Fig. \ref{fig:how} and detailed next.

\subsection{Architecture}


\textbf{Sentence vectorization}.
%
To handle sentences of varied length, we choose to first vectorize each sentence. In particular, we consider three common text vectorization strategies, \ie, bag-of-words, word hashing and \wordvec, in the order of increasing model scalability.

\textbf{\textit{Strategy I: Bag-of-words}}. 
Bag-of-words is a classical representation used in text analysis. Each dimension in a bow vector corresponds to the occurrence of a specific word in the input text, \ie,
\begin{equation}
s_{bow}(q) = (c(w_1,q), c(w_2,q),\ldots,c(w_m,q)),
\end{equation}
where $c(w,q)$ returns the occurrence of word $w$ in $q$, and $m$ is the size of the vocabulary. While previous works keep words occurring at least five times in the training set to ensure meaningful probabilistic estimation \cite{cvpr2015-neuraltalk,vinyals2015show}, we avoid this heuristic rule, preserving all words in the training data.

\textbf{\textit{Strategy II: Word hashing}}.
Due to the use of letter based $n$-grams, word hashing has the advantage of reducing the size of the input layer while generalizing well to infrequent and unseen words \cite{huang2013learning}, when compared to bag-of-words. Following \cite{fang2015captions}, we decompose each word in a given text into a list of letter-trigrams, \eg, \emph{cat} as \{\#ca, cat, at\#\}, and then count their occurrence to obtain a letter-trigram count vector $s_{hashing}(q)$.
Considering the Flickr30k corpus for instance, the word hashing vectorization reduces the size of the input layer from 17,723 to 5,969.

\textbf{\textit{Strategy III: \wordvec}}.
Let $w2v(w)$ be individual word embedding vectors, we obtain the embedding vector of the input text by mean pooling over its words, \ie,
\begin{equation}
s_{word2vec}(q) :=\frac{1}{|q|}\sum_{w \in q} w2v(w),
\end{equation}
where 
$|q|$ is the sentence length. Previous works employ \wordvec~trained on web documents as their word embedding matrix \cite{nips13devise,vinyals2015show,iccv15-huawei}. However, recent studies suggest that \wordvec~trained on Flickr tags better captures visual relationships than its counterpart learned from web documents \cite{li2015zero}. We therefore train by the skip-gram algorithm \cite{word2vec} a 500-dimensional model using English tags of 30 million Flickr images. The use of \wordvec~reduces the size of the input layer to 500 only, meanwhile it supports a much larger vocabulary of 1.7 million words.

\textbf{Text transformation via a multilayer perceptron}. 
The output of the first layer $s(q)$ goes through subsequent hidden layers until it reaches the output layer $r(q)$, which resides in the visual feature space.
More concretely, by applying an affine transformation on $s(q)$, followed by an element-wise ReLU activation $\sigma(z) = \max(0,z)$,
we obtain the first hidden layer $h_1(q)$ of an $l$-layer \ourmodel~as: 
\begin{equation}
h_1(q)=\sigma(W_{1}  s(q)+b_{1}).
\end{equation}
The following hidden layers are expressed by:
\begin{equation}
h_i(q)=\sigma(W_{i} h_{i-1}(q)+b_{i}), i = 2,...,l-2,
\end{equation}
where $W_{i}$ parameterizes the affine transformation of the $i$-th hidden layer and $b_{i}$ is a bias terms.
In a similar manner we compute the output layer $r(q)$ as: 
\begin{equation}
r(q)=\sigma(W_l  h_{l-1}(q)+b_l).
\end{equation}
Putting it all together, the learnable parameters are represented by $\theta=[W_1, b_1, \ldots, W_l, b_l]$.
                                                         
In principle the learning capacity of our model grows as more layers are used. This also means more solutions exist which minimize the training loss, yet are suboptimal for unseen test data. We study in Section \ref{ssec:eval-design-choice} how deep \ourmodel~can go without losing generalization ability.

\subsection{Learning algorithm}

\textbf{Objective function}. 
While an image is worth a thousand words, a sentence is meant to describe the main objects and scene in the image.
This is in a way similar to the visual feature extracted by a ConvNet, trained to capture the essentials of the pictorial content. In that regard, given an image $x$ and a sentence $q$ describing the image, 
we propose to reconstruct its visual feature $\phi(x)$ directly from $q$, with Mean Squared Error (MSE) as our objective function. We have also experimented with the marginal ranking loss, as commonly used in previous works \cite{grangier2008discriminative,Bai2009Polynomial,nips13devise,cvpr2015-neuraltalk}, but found MSE yields better performance.

The MSE loss $l_{mse}$ for a given training pair is defined as: 
\begin{equation} \label{eq:loss_mse}
l_{mse}(x,q ;\theta) =  (r(q)-\phi(x))^2.
\end{equation}
We train \ourmodel~to minimize the overall MSE loss on a given training set $\mathcal{D} = \{(x,q)\}$, containing a number of relevant image-sentence pairs:
\begin{equation}\label{eq:obj_mse}
\argmin_{\theta} \sum_{(x,q)\in \mathcal{D}}  l_{mse}(x,q ;\theta).
\end{equation}

For a given image, different persons might describe the same visual content in different words. \emph{E.g.},  ``A dog leaps over a log'' versus ``A dog is leaping over a fallen tree''. The verb leap in different tenses essentially describe the same action, while a log and a fallen tree can have similar visual appearance. Projecting the two sentences into the same visual feature space has the effect of implicitly finding such correlations. Indeed, as shown in Section \ref{ssec:text-to-text}, for text retrieval in image sentence corpora \ourmodel~is found to be a better representation of text than \wordvec.

\textbf{Optimization}.
We solve Eq. (\ref{eq:obj_mse}) using stochastic gradient descent with RMSprop \cite{tijmen2012rmsprop}.
This optimization algorithm 
%
%
divides the learning rate by an exponentially decaying average of squared gradients, to prevent the learning rate from effectively shrinking over time.
We empirically set the initial learning rate $\eta=0.001$, decay weights $\gamma=0.9$ and small constant $\epsilon=10^{-06}$ for RMSprop.
In addition, we apply dropout to the hidden layer\cite{nips2012-hinton}. 
%
Lastly, we apply an early stop strategy: stop training if there is no performance improvement on the validation set in 5 successive epochs, with the maximal number of epochs set to be 500.


\subsection{Image to sentence}
For a given image, we tackle the image-to-sentence task by selecting from the sentence pool the one deemed most relevant with respect to the image. This is achieved by sorting all the sentences in light of a specific similarity $sim(x,q)$. For each sentence $q$, we obtain its visual encoding $r(q)$ by forward computation through the \ourmodel~network. Subsequently the similarity between the sentence and the image is computed as the cosine similarity between $r(q)$ and the image feature $\phi(x)$:
\begin{equation} \label{eq:simi_fun}
sim(x,q) :=\frac{r(q) \cdot \phi(x) }{\left \| r(q)  \right \| \left \| \phi(x) \right \|}
\end{equation}
We choose this similarity as it normalizes feature vectors and is found to be better than the dot product.


\subsection{Video to sentence} \label{sec:w2vv-video}

\ourmodel~is also applicable in the video domain as long as we have an effective vectorized representation of video. Again, different from previous video-to-sentence models that execute in a joint subspace \cite{aaai2015-xu-video,eccv2016ws-otani}, we project sentences into the video feature space.

Following the good practice of using pre-trained ConvNets for video content analysis \cite{naacl2015-video-to-text,ye2015,mettes2016imagenet,pan2016jointly}, we extract features by applying image ConvNets on individual frames and 3-D ConvNets \cite{iccv2015-c3d} on consecutive-frame sequences.
For the short video clips we experiment with, mean pooling features over video frames is considered reasonable \cite{naacl2015-video-to-text,pan2016jointly,cvpr2016-yu-video}. 
Hence, we use the mean pooling strategy to obtain a visual feature vector per video.

The audio channel of a video can sometimes  provide complementary information to the visual channel. For instance, to help decide whether a person is talking or singing. To exploit this channel, we extract a bag of quantized Mel-frequency Cepstral Coefficients (MFCC) vector \cite{opensmile} and concatenate it with the previous visual feature.
\ourmodel~is trained to predict such a visual-audio feature, as a whole, from input text.

\ourmodel~is used in a principled manner, transforming an input sentence to a video feature vector, let it be visual or visual-audio. For the sake of clarity we term the video variant \textit{Word2VideoVec}.

\section{Experiments} \label{sec:eval}


\subsection{Properties of Word2VisualVec} \label{ssec:eval-design-choice}

We first investigate the impact of major design choices on \ourmodel. Due to high complexity of the problem, evaluating all the variables simultaneously is computationally prohibitive. The evaluation is thus conducted sequentially, focusing on one variable per time. For its efficient execution, \wordvec~is used as the first layer.

\textbf{Data}.  
We use Flickr8k \cite{flickr8k} and Flickr30k \cite{flickr30k}, two popular benchmark sets for the image-to-sentence task.
Each image is associated with five crowd-sourced English sentences, which briefly describe the main objects and scenes present in the image. For the ease of cross-paper comparison, we follow the identical data partitions as used in \cite{cvpr2015-neuraltalk,iclr15-ma-mrnn}. 
That is, training / validation / test is 6k / 1k / 1k for Flickr8k and 29,783 / 1k / 1k for Flickr30k.

\textbf{Evaluation criteria}.
Following the common convention \cite{flickr8k,cvpr2015-neuraltalk,iclr15-ma-mrnn,iccv15-huawei,iccv15-flickr30kentities}, we report rank-based performance metrics $R@K$ ($K = 1, 5, 10$) and Median rank ($Med~r$). $R@K$ computes the percentage of test images for which at least one correct result is found among the  top-$K$ retrieved sentences, and $Med~r$ is the median rank of the first correct result in the ranking. Hence, higher $R@K$ and lower $Med~r$ means better performance.
Due to space limit we report results on Flickr8k. Similar results have been observed on Flickr30k, see the supplementary material.

\textbf{Which visual feature?} 
A deep visual feature is determined by a specific ConvNet and its layers. We experiment with four pre-trained ConvNets, \ie, CaffeNet \cite{Jia2014Caffe}, GoogLeNet \cite{googlenet}, ResNet-152 \cite{cvpr2016-resnet}, and GoogLeNet-shuffle \cite{mettes2016imagenet}. The first three ConvNets were trained using images containing 1K different visual objects as defined in the Large Scale Visual Recognition Challenge \cite{ILSVRCarxiv14}. GoogLeNet-shuffle follows GoogLeNet's architecture, but is re-trained using a bottom-up reorganization of the complete 22K ImageNet hierarchy, excluding over-specific classes and classes with few images and thus making the final classes have balanced positive images. We tried multiple layers of each ConvNet model and report the best performing layer. As shown in Table \ref{tab:visual_feat}, as the ConvNet goes deeper, predicting the corresponding visual features by \ourmodel~improves. This result is encouraging as better performance can be expected from the continuous progress in deep learning features. In what follows we rely on the ResNet-152 feature because of its top performance.


\begin{table} [tb!]
\renewcommand{\arraystretch}{1.2}
\caption{\textbf{Which visual feature?}. For all the features, we use a three-layer Word2VisualVec. 
Predicting the ResNet-152 feature yields the best performance.}
\label{tab:visual_feat}
\centering
 \scalebox{0.8}{
\begin{tabular}{l l  rrrr}
\toprule
\textbf{ConvNet} & \textbf{Layer}  & R@1 & R@5 & R@10 & Med r \\
\cmidrule(l){1-6}
CaffeNet        & fc7      & 19.6  & 42.2 &   54.7  &  8 \\
GoogLeNet       & pool5    & 24.7  & 51.7 &   62.7  &  5 \\
GoogLeNet-shuffle & pool5  & 30.1  & 57.9 &   70.7  &  4 \\
ResNet-152        & pool5  & \textbf{31.9} &  \textbf{61.8} & \textbf{75.3} & \textbf{3} \\
\bottomrule
\end{tabular}
 }
\end{table}

\begin{table} [tb!]
\renewcommand{\arraystretch}{1.2}
\caption{\textbf{How deep?}. Image-to-sentence results on Flickr8k indicate a three-layer Word2VisualVec, \ie, 500-1000-2048, strikes the best balance between model capacity and generalization ability.
}
\label{tab:model_depth}
\centering 
 \scalebox{0.78}{
\begin{tabular}{r l r r r r}
\toprule
\textbf{Layers}& \textbf{Net architecture}  & R@1 & R@5 & R@10 & Med r \\
\cmidrule(l){1-6}
2 & 500-2048                 & 28.3 & 55.1 & 68.5 & 4 \\
3 & 500-1000-2048            & \textbf{31.9} & \textbf{61.8} & \textbf{75.3} & \textbf{3} \\
4 & 500-1000-1000-2048       & 29.9 & 60.6 & 72.0 & \textbf{3} \\
5 & 500-1000-1000-1000-2048  & 27.3 & 53.5 & 68.0 & 5 \\
\bottomrule
\end{tabular}
 }
\end{table}

\begin{table*} [tb!]
\renewcommand{\arraystretch}{1.2}
\caption{\textbf{How to vectorize an input sentence?}. Image-to-sentence results given distinct input layers, where $\rightarrow$ indicates a cross-dataset scenario, \eg, testing models derived from Flickr30k on the Flickr8k test set or vice versa. Vectorizing input sentences by bag-of-words provides the best overall performance, while using word hashing or word2vec are slightly better when training data is limited.}
\label{tab:exp-first-layer}
\centering 
\scalebox{0.8}{
\begin{tabular}{@{}l rrrr rrrr | rrrr rrrr@{}}
\toprule
& \multicolumn{4}{c}{\textbf{Flickr8k}} & \multicolumn{4}{c}{\textbf{Flickr30k}} & \multicolumn{4}{c}{\textbf{Flickr30k $\rightarrow$ Flickr8k}} & \multicolumn{4}{c}{\textbf{Flickr8k $\rightarrow$ Flickr30k}} \\
\cmidrule(l){2-5}  \cmidrule(l){6-9} \cmidrule(l){10-13} \cmidrule(l){14-17}
\textbf{Input layer}  & R@1 & R@5 & R@10 & Med r & R@1 & R@5 & R@10 & Med r & R@1 & R@5 & R@10 & Med r & R@1 & R@5 & R@10 & Med r\\
\cmidrule{1-17}
bag-of-words  & 33.6  & \textbf{62.0} & \textbf{75.3}  & 3 & 39.7  & \textbf{67.0}  &  \textbf{76.7} &  \textbf{2} & \textbf{40.3}  & \textbf{69.3} & \textbf{81.0} & \textbf{2}  &  26.7  & 50.7 & 60.7  & 5\\
word hashing  & \textbf{33.9} & 62.0 & 73.4 & 3 & \textbf{40.3}  &  65.6  & 73.3  & 2  &  37.4  & 65.3 & 77.3 & 3 &  27.5   &  51.1  & 62.7  & 5 \\
word2vec & 31.9 & 61.8 & 75.3 & 3 & 36.6  & 64.2  & 74.2  & \textbf{2} & 34.8 & 63.8 & 76.5 & 3 &   \textbf{27.9}  &  \textbf{54.7} & \textbf{66.2}  & 5\\ 
%
\bottomrule
\end{tabular}
 }
\end{table*}

\textbf{How deep?} 
Table \ref{tab:model_depth} shows the performance of \ourmodel~as the number of its layers increases. Note that the 500-1024 architecture resembles the model from Frome \etal \cite{nips13devise} but with the mapping direction reversed. As more layers are in use, \ourmodel~improves, at the cost of an increased learning complexity. On both Flickr8k and Flickr30 the performance peak is reached for three layers, \ie, 500-1000-2048. We observe performance degeneration as the model depth goes beyond three. Recall that the model is chosen in terms of its performance on the validation set. While its learning capacity increases as the model goes deeper, the chance of overfitting also increases. The three-layer \ourmodel~strikes the best balance between model capacity and generalization ability.

\textbf{How to vectorize an input sentence?}
So far we have used word2vec as the first layer to study the main properties of \ourmodel. Here we investigate the impact of alternative vectorization choices for the first layer. When considering Flickr8k and Flickr30k individually (Table \ref{tab:exp-first-layer}), bag-of-words performs the best. This is not surprising as word hashing and word2vec trade performance for higher efficiency in the training stage. We also consider a cross-dataset experiment, where we apply models trained on Flickr30k to the Flickr8k test set, and vice versa (Table \ref{tab:exp-first-layer}). For the Flickr30k $\rightarrow$ Flickr8k transfer, bag-of-words is still the best choice, but for the other way around both word hashing and word2vec are better, indicating their better generalization ability when training data is limited.


\textbf{Data sensitivity?}
To further study the influence of the amount of training images on \ourmodel, we generate training subsets of size 1k, 2k, and 4k, by randomly down-sampling the Flickr8k training data. Per size the procedure repeats five times. The performance mostly converges when only half of the training data is exploited. Note that less training data means a higher chance of encountering unseen words in the test stage. Sentence vectorization by word2vec alleviates the issue best, as it outperforms bag of words and word hashing when learning from only 1k and 2k images, see Fig. 1 in the supplementary material.

\textbf{How fast?}
We implement \ourmodel~using Keras\footnote{\url{https://github.com/fchollet/keras}}, a deep learning library written in Python. 
Per training epoch \ourmodel~with \wordvec~based vectorization takes the least execution time. 
However, the model with word hashing needs less number of training epochs. 
For the three-layer model with its input layer instantiated as bag of words, word hashing, and \wordvec, 
it takes in total 8, 3.2, and 6.5 hours to learn from Flickr30k of 149k image-sentence pairs on a GeoForce GTX 1070 GPU. 
Prediction is swift at an averaged speed of 0.86 millisecond per forward propagation.


Based on the above evaluations we recommend a three-layer \ourmodel~that predicts the ResNet-152 feature,
and uses bag-of-words for the input layer given adequate training data and \wordvec~for learning from scarce resources.
We will release code and models.

\subsection{Potential of Visual Space for Text Retrieval}

While \ourmodel~is meant for image to sentence by projecting a given sentence into the visual space, it essentially generates a new representation of text. How meaningful is this new representation as compared to existing ones in the text field such as classical bag-of-words and the recent word2vec? 
%
%
This experiment conducts text-only retrieval, where candidate sentences are sorted in descending order according to their relevance scores to a given query sentence. We predict visual features for both the input query sentence and the text in the test set. 
Again, the cosine distance is used to measure pairwise similarity, and subsequently used to generate sentence rankings. 

\textbf{Setup}.
We need pairs of sentences that are visually and semantically relevant. Textual descriptions in Flickr8k and Flickr30k meet this requirement as they are meant for describing the same visual content. Moreover, since they were independently written by distinct users, the wording may vary across the users, requiring a text representation to capture shared semantics among distinct words. We construct two test sets for text retrieval using the Flickr8k and Flickr30k test sets, respectively. Per set, we treat the first sentence of a test image as a query and the other four sentences as test instances, resulting in a total of 1k queries and a test pool of 4k sentences.
Every sentence is vectorized by bag-of-words, 
mean pooling of \wordvec, 
and \ourmodel, respectively. Mean Average Precision (mAP) is reported. 





\textbf{Results}. 
On both test sets \ourmodel~scores higher mAP than bag-of-words and \wordvec, see Table \ref{tab:text-to-text}.
We notice that for both bag-of-words and word2vec, excluding stop words in advance gives a clear boost in performance,
showing the importance of preprocessing for the two representations. By contrast, the contribution of the same preprocessing to \ourmodel~is relatively limited, as the stop word effect has been minimized during the learning process. While the visual space has been extensively exploited for vision tasks, this experiment reveals its potential for performing text only retrieval as well.


\begin{table} [tb!]
\renewcommand{\arraystretch}{1.2}
\caption{\textbf{Potential of visual space for text retrieval}. The \ourmodel~model is directly taken from the previous experiment without re-training. For finding text in image sentence corpora, the visual space is more suited than the classical bag-of-words space and the recent word2vec space.} 
\label{tab:text-to-text}
\centering 
\scalebox{0.8}{
\begin{tabular}{@{} l  r r @{}}
\toprule
\textbf{Text representation}  & \textbf{Flickr8k} & \textbf{Flickr30k} \\
\cmidrule{1-3}
bag-of-words                     			 & 16.2  & 16.0 \\ 
bag-of-words, stop words excluded from input & 26.3  & 28.1 \\
word2vec                         			 & 18.3  & 14.7 \\ 
word2vec, stop words excluded from input   	 & 31.1  & 28.4 \\ [3pt]
\ourmodelbow                     			 & 33.1  & \textbf{40.0} \\
\ourmodelwv                      			 & \textbf{36.2} &   34.1 \\
\bottomrule
\end{tabular}
 }
\end{table}
 \label{ssec:text-to-text}

\subsection{Video-to-Sentence}

\textbf{Setup}. 
We compare with the joint embedding by Xu \etal \cite{aaai2015-xu-video}, its improved version by Otani \etal\cite{eccv2016ws-otani}, and LSTM-E from Pan \etal \cite{pan2016jointly}.
For fair comparison to \cite{aaai2015-xu-video,pan2016jointly}, we follow their evaluation protocol, 
reporting $Mean~r$ on the Microsoft Video Description dataset (MSVD) \cite{chen2011collecting}, with 1,200, 100 and 670 video clips for training, validation, and test. 
Otani \etal \cite{eccv2016ws-otani} also use MSVD, but has the test set down-sampled by randomly choosing 5 sentences per test video.


\textbf{Results}. As Table \ref{tab:exp-vtt} shows, \videovec~outperforms all the competitor models. As the MSVD videos were muted, we cannot evaluate \videovec~with audio.
Among the three visual features, ResNet-152 performs the best, followed by GoogLeNet-shuffle and C3D.
Given the same feature, using \wordvec~as the input layer is better than bag-of-words. 
This is in line with our conclusion in Section \ref{ssec:eval-design-choice} that vectorizing input sentences by \wordvec~is preferred given relatively limited amounts of training data.

\begin{table} [tb!]
\renewcommand{\arraystretch}{1.2}
\caption{\textbf{Video-to-sentence results on MSVD}.  All numbers are from the cited papers.
\videovec,~with \wordvec~as the input layer and predicting the ResNet-152 feature, scores the best.}
\label{tab:exp-vtt}
\centering 
 \scalebox{0.74}{
\begin{tabular}{@{}l  r r r r@{}}
\toprule
\textbf{Model}  & R@1 & R@10 & Med r  & Mean r \\
\cmidrule{1-5}
\multicolumn{5}{c}{\textbf{Results using data partition from Xu \etal}\cite{aaai2015-xu-video} } \\
Xu \etal \cite{aaai2015-xu-video} & -- & -- & -- & 224.1 \\
Pan \etal \cite{pan2016jointly}   & -- & -- & -- & 208.5 \\ 
\videovecbow $\Rightarrow$ C3D               & 9.7  &  26.4 &  64 & 358.0 \\
\videovecwv $\Rightarrow$ C3D                & 11.6 &  34.9 &  35 & 248.2 \\
\videovecbow $\Rightarrow$ GoogLeNet-shuffle & 14.2 &  36.7 &  30 & 220.6\\
\videovecwv $\Rightarrow$ GoogLeNet-shuffle  & \textbf{16.4} &  40.3 &  17 & 148.6 \\
\videovecbow $\Rightarrow$ ResNet-152        & 15.7 &  41.5 &  21 & 171.9 \\
\videovecwv $\Rightarrow$ ResNet-152         & 16.3 & \textbf{44.8} &  \textbf{14} & \textbf{110.2} \\

\multicolumn{5}{c}{\textbf{Results using data partition from Otani \etal}\cite{eccv2016ws-otani} } \\
Otani \etal\cite{eccv2016ws-otani}    & 9.9 & 38.4 & 19 & 75.2 \\ 
\videovecwv $\Rightarrow$ ResNet-152         & \textbf{17.9} & \textbf{49.4} &  \textbf{11} & \textbf{57.6}\\
\bottomrule
\end{tabular}
 }
\end{table}

\subsection{Comparison to the State-of-the-Art}  \label{ssec:image-to-text}

\subsubsection{Image to Sentence}

\textbf{Setup}.
We compare a number of recently developed image-to-sentence models including deep visual-embedding by Frome \etal \cite{nips13devise},
stacked auxiliary embedding by Gong \etal \cite{eccv2014-sentence-embedding},
bidirectional RNN by Karpathy \& Fei-Fei \cite{cvpr2015-neuraltalk},
neural image caption by Vinyals \etal \cite{vinyals2015show},
multimodal RNN by Mao \etal \cite{iclr15-ma-mrnn},
multimodal CNN by Ma \etal \cite{iccv15-huawei},
bidirectional LSTM by Wang \etal \cite{mm16-bilstm},
linear CCA with fisher vector pooling over word vectors by Klein \etal \cite{cvpr2015-klein-fv},
and region KCCA by Plummer \etal \cite{iccv15-flickr30kentities}.

\textbf{Results}.
Table \ref{tab:image_to_sent_perf} presents the performance of the above models on both Flickr8k and Flickr30k.
\ourmodel~compares favorably against the state-of-the-art.
Notice that Plummer \etal \cite{iccv15-flickr30kentities} employ extra bounding-box level annotations.
Still our results are slightly better.
As all the competitor models use joint subspaces,
the results justify the viability of directly using deep visual feature space for image-to-sentence matching.

%
%


\begin{table} [tb!]
\renewcommand{\arraystretch}{1.2}
\caption{\textbf{State-of-the-art image-to-sentence results}. 
All numbers are from  the cited papers except for Frome \etal \cite{nips13devise} which is based on our implementation.
\ourmodel~outperforms recent alternatives, even when they use additional region annotations~\cite{iccv15-flickr30kentities}, showing the effectiveness of deep visual feature space for image-to-sentence matching.}
\label{tab:image_to_sent_perf}
\centering 
\scalebox{0.75}{
\begin{tabular}{@{}l  r rr  rrr @{}}
\toprule

 & \multicolumn{3}{c}{\textbf{Flickr8k}} & \multicolumn{3}{c}{\textbf{Flickr30k}} \\
\cmidrule{2-4}  \cmidrule{5-7}
  & R@1 & R@10 & Med r & R@1 & R@10  & Med r\\
\cmidrule{1-7}
Frome \etal \cite{nips13devise} & 11.9 & 45.8  & 13 & 12.1 & 48.6 & 11\\
Gong \etal \cite{eccv2014-sentence-embedding}  &  --   & 48.8    &  --      &  -- & -- & --\\
Karpathy \& Fei-Fei \cite{cvpr2015-neuraltalk}  & 16.5 & 54.2 & 7.6         &  22.2   & 61.4 & 4.8 \\
Vinyals \etal \cite{vinyals2015show}            & 20.0 & 61.0 & 6           &  17.0   & 56.0 & 7 \\ 
Mao \etal \cite{iclr15-ma-mrnn}                 & 14.5 & 48.5 & 11          &  35.4   & 73.7 & 3 \\ 
Ma \etal \cite{iccv15-huawei}                   & 24.8 & 67.1 & 5           &  33.6   & 74.9 & 3 \\
Wang \etal \cite{mm16-bilstm}                   & 29.3 & 69.6 & \textbf{3}  &  28.1   & 64.2 & 4 \\
Klein \etal \cite{cvpr2015-klein-fv}            & 31.0 & 73.7 & 4           &  33.3   & 74.7 & 3 \\
Plummer \etal \cite{iccv15-flickr30kentities}   & -- & --  & --   &  39.1   & 76.4 & -- \\ [3pt]
\ourmodel & \textbf{33.6}  &  \textbf{75.3}  &  \textbf{3}  & \textbf{39.7} &  \textbf{76.7} &   \textbf{2} \\
\bottomrule
\end{tabular}
 }
\end{table}


\subsubsection{Video to Sentence}

\begin{figure*}[tb!]
\centering
 \subfigure[Results on the given set A]{
\noindent\includegraphics[width=0.82\columnwidth]{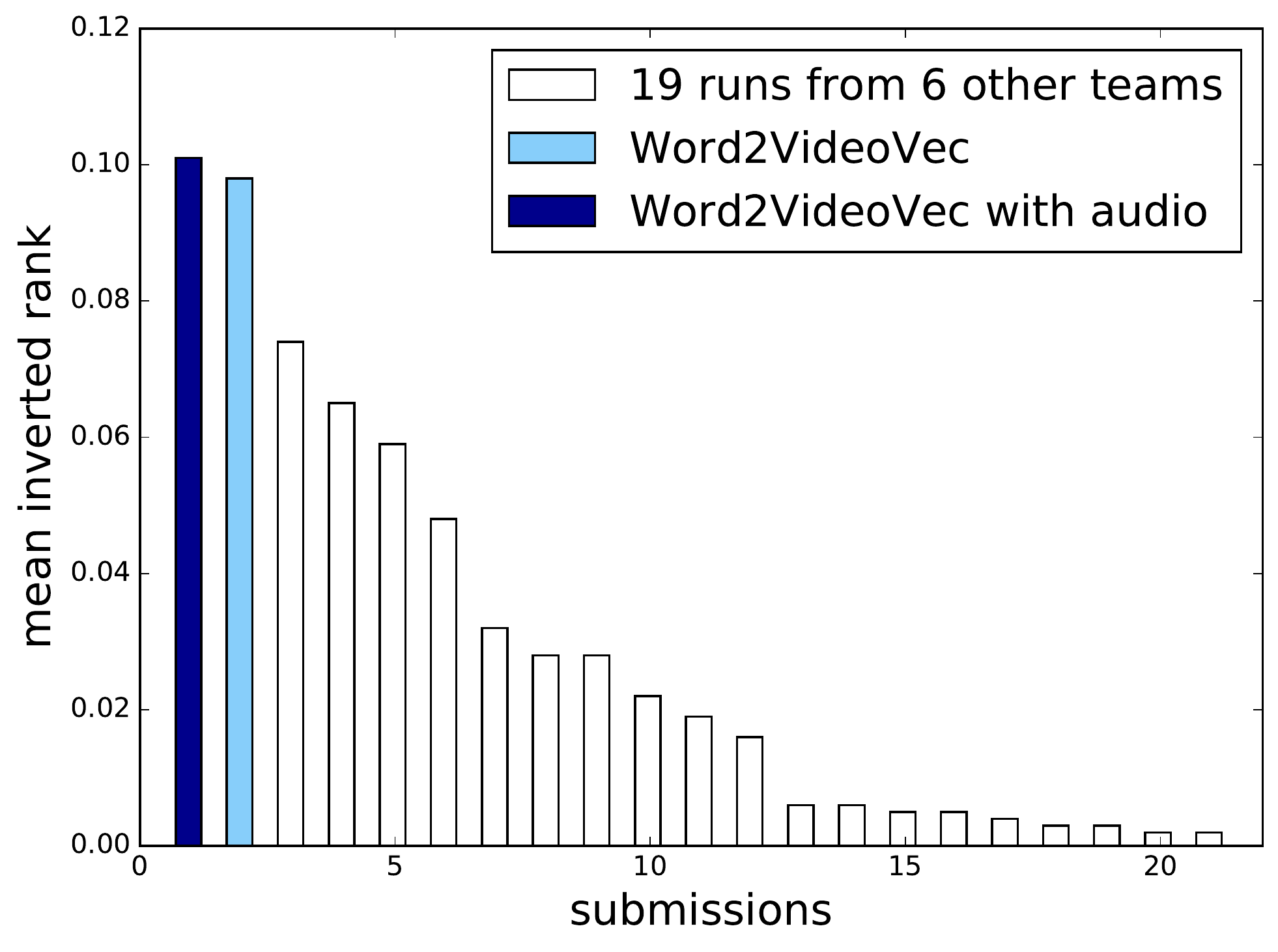}
\label{fig:tv16-setA}
}
 \subfigure[Results on the given set B]{
\noindent\includegraphics[width=0.82\columnwidth]{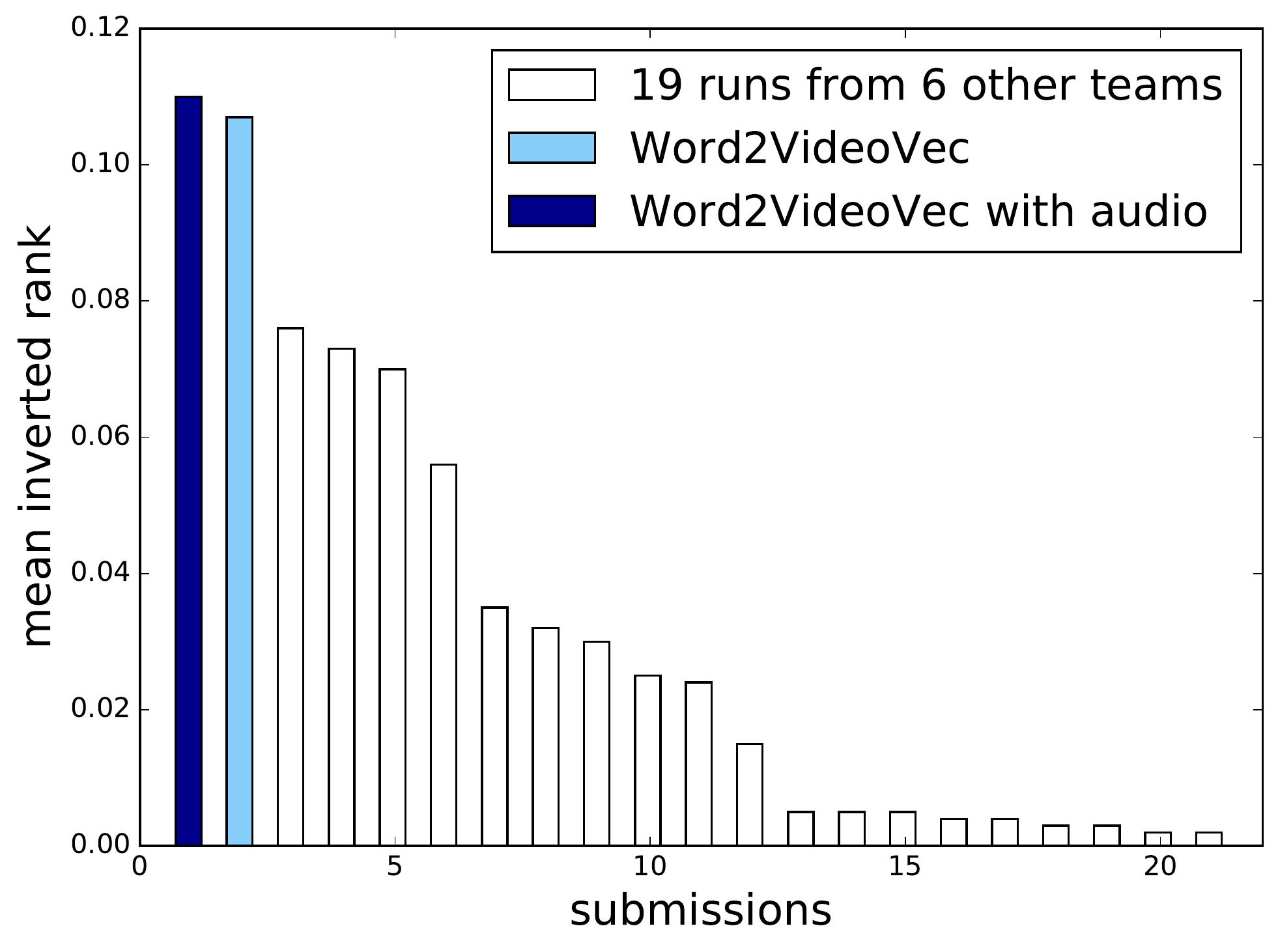}
\label{fig:tv16-setB}
}
\caption{
\textbf{State-of-the-art video-to-sentence results} in the TrecVid 2016 benchmark, showing the good performance of \videovec~compared to 19 alternative approaches,
which can be further improved by predicting the visual-audio feature.}
 \label{fig:trecvid_submission}
\end{figure*}

\begin{figure*}[tb!]
\centering\includegraphics[width=2\columnwidth]{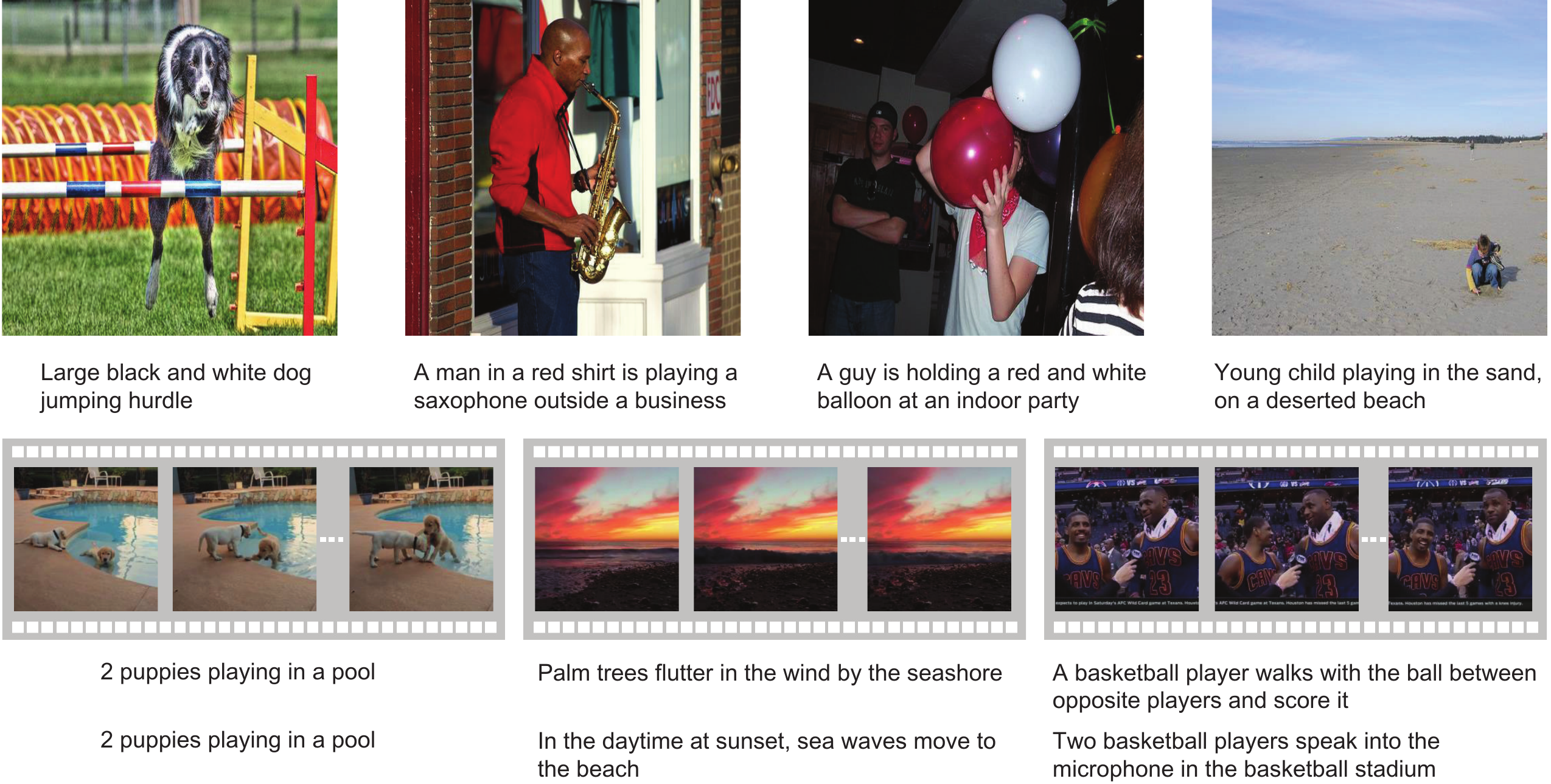}
\caption{Some vision to language matching results by this work.
The last row are the sentences matched by \videovec~with audio, showing that adding audio sometimes help describe acoustics, e.g. \textit{sea wave} and \textit{speak}. 
See more results in the supplementary material.
}\label{fig:caption-results}
\end{figure*}

\textbf{Setup}.
We also participated in the NIST TrecVid 2016 Video-to-Text matching and ranking task~\cite{AwadTRECVID16}. In this task, participants were asked to rank a list of pre-defined sentences in terms of relevance for a given video. The test set consists of 1,915 videos collected from Twitter Vine. Each video is about 6 sec long. The videos were given to 8 annotators to generate a total of 3,830 sentences, with each video associated with two sentences written by two different annotators. The sentences have been split into two equal-sized subsets, set $A$ and set $B$, with the rule that sentences describing the same video are not in the same subset. Per test video, participants are asked to rank all sentences in the two subsets. Notice that we have no access to the ground-truth, as the test set is used for blind testing by the organizers only. NIST also provides a training set of 200 videos, which we consider insufficient for training \videovec. Instead, we learn the network parameters using video-text pairs from MSR-VTT \cite{xu2016msr}, with hyper-parameters tuned on the provided TrecVid training set. By the time of TrecVid submission, we used GoogLeNet-shuffle as the visual feature, a 1,024-dim bag of MFCC as the audio feature, and \wordvec~as the input layer. The performance metric is Mean Inverted Rank at which the annotated item is found. 
Higher mean inverted rank means better performance.

\textbf{Results}. 
As shown in Fig. \ref{fig:trecvid_submission},  with Mean Inverted Rank ranging from 0.097 to 0.110, \videovec~leads the evaluation on both set A and set B in the context of all submissions from seven teams worldwide. 
Moreover, the results can be further improved by predicting the visual-audio feature. 

Some qualitative matching results are shown in Fig. \ref{fig:caption-results}. 

\section{Conclusions} \label{sec:conc}

For image to sentence matching, we advocate deep visual features as a  new shared representation of images and sentences. \ourmodel~is proposed to project vectorized sentences into a given visual feature space, be it a GoogleNet or a ResNet. We show how the model can be generalized to the video domain by predicting a prescribed video feature, including audio, again from sentences. We obtain state-of-the-art results on Flickr8k and Flickr30k for image to sentence matching, 
and MSVD and the TrecVid 2016 Video-to-Text benchmark for video to sentence matching. What is more we also demonstrate the potential of the visual space for text retrieval.

{\small

}

\end{document}